\documentclass[pdflatex,sn-mathphys-num]{sn-jnl}

\usepackage{graphicx}%
\usepackage{multirow}%
\usepackage{amsmath,amssymb,amsfonts}%
\usepackage{amsthm}%
\usepackage{mathrsfs}%
\usepackage[title]{appendix}%
\usepackage{xcolor}%
\usepackage{textcomp}%
\usepackage{manyfoot}%
\usepackage{booktabs}%
\usepackage{algorithm}%
\usepackage{algorithmicx}%
\usepackage{algpseudocode}%
\usepackage{listings}%
\usepackage{pdflscape}
\DeclareMathOperator*{\argmax}{arg\,max}

\theoremstyle{thmstyleone}%
\theoremstyle{thmstyletwo}%

\theoremstyle{thmstylethree}%

\begin{document}

\title[Cross-Entropy Attacks to Language Models via Rare Event Simulation]{Cross-Entropy Attacks to Language Models via Rare Event Simulation}


\author[1]{\fnm{Mingze} \sur{Ni}}\email{mingze.ni@uts.edu.au}

\author[2]{\fnm{Yongshun} \sur{Gong}}\email{ysgong@sdu.edu.cn}

\author*[1]{\fnm{Wei} \sur{Liu}}\email{wei.liu@uts.edu.au}

\affil*[1]{\orgdiv{School of Computer Science}, \orgname{University of Technology Sydney}, \orgaddress{\street{15 Broadway Ultimo}, \city{Sydney}, \postcode{2007}, \state{NSW}, \country{Australia}}}

\affil*[2]{\orgdiv{School of Software}, \orgname{Shandong University}, \orgaddress{\street{1500 Shunhua Road}, \city{Jinan}, \postcode{250000}, \state{Shandong}, \country{China}}}



\abstract{Black-box textual adversarial attacks are challenging due to the lack of model information and the discrete, non-differentiable nature of text. Existing methods often lack versatility for attacking different models, suffer from limited attacking performance due to the inefficient optimization with word saliency ranking, and frequently sacrifice semantic integrity to achieve better attack outcomes. This paper introduces a novel approach to textual adversarial attacks, which we call Cross-Entropy Attacks (CEA), that uses Cross-Entropy optimization to address the above issues. Our CEA approach defines adversarial objectives for both soft-label and hard-label settings and employs CE optimization to identify optimal replacements. Through extensive experiments on document classification and language translation problems, we demonstrate that our attack method excels in terms of attacking performance, imperceptibility, and sentence quality.
}



\maketitle
\section{Introduction}
Deep neural networks (DNNs) have gone through rapid development in recent years, and they have been successfully utilized in various natural language processing (NLP) tasks such as text classification and natural language inference. However, NLP models are known to be vulnerable across various applications, including machine translation \cite{ni2022attacking, cheng2020seq2sick, morphin2020tan}, sentiment analysis \citep{zang2020word, yang2021bigram}, and text summarization \citep{cheng2020seq2sick}. Attackers can exploit these weaknesses to create adversarial examples that compromise the performance of targeted NLP systems, posing significant security challenges.

Attacks to textual classifiers are commonly with two primary black-box settings: the {soft-label} and the {hard-label} setting. The soft-label setting, offering access to the predicted label and confidence score, enables attackers to significantly deceive victim models using customized word saliency \cite{zang2020word, li2021clare, li2020bertattack}. However, this approach is often impractical due to limited access to confidence scores. In the contrast, the hard-label setting, where only the predicted label is available, is more challenging and realistic, leading researchers to use customized optimization techniques to perturb the original text and compromise victim models \cite{ye2022texthoaxer, liu2023sspattack, zhu2024limeattack}, though there is still room for improvement in their effectiveness. In addition to classifiers, researchers have extensively explored attacks on sequence-to-sequence models, typically leveraging word importance rankings and optimizing targeted pre-selected words. Notable methods, such as those based on attention scores and BLEU score drops, have proven effective in attacking neural machine translation (NMT) models \cite{ni2022attacking, morphin2020tan}. 

Despite the success of these methods, several issues remain. No existing literature offers a unified framework for attacking both classifiers and sequence-to-sequence models, and cross-domain adaptations often result in suboptimal performance, highlighting the limited versatility of current approaches \cite{ni2022attacking}. We argue that adversarial attacks on NLP models should be consistent across model types, given their shared characteristics, such as discreteness, sequential structure, and attention-based learning mechanisms \cite{attention}. Additionally, soft-label settings heavily rely on word importance ranks \cite{zang2020word, li2020bertattack, li2021clare, yang2021bigram}, which have drawbacks such as ineffective ranking metrics and lack of access to confidence scores, making them less practical than hard-label attacks. Hard-label attacks, while increasingly popular, still face high optimization costs due to large word embedding spaces \cite{hardfirst, ye2022texthoaxer} and inefficient word importance ranking \cite{liu2023sspattack, zhu2024limeattack}, affecting attack performance. For attacking NMTs, most methods are inherited from classifier attacks and have significant potential for performance improvements.

To address these issues,  we first define common objectives for both classifiers and NMT, then propose a cross-entropy-based (CE) algorithm to optimize the attack objectives based on rare event simulations. Specifically, the CE algorithm simulates events with probabilities exceeding a high threshold, which is iteratively increased for optimization. This approach aligns with the logic of posing adversarial attacks by treating adversarial examples as rare events and the objective as the threshold. 

Our main contributions are as follows:
\begin{itemize}
    \item We have designed a highly effective adversarial attack method called the Cross-Entropy Attack (CEA), utilizing rare event simulation. This approach innovatively generates attacks by treating adversarial examples as rare errors in well-trained NLP models. 
    \item Our proposed attack method is applicable not only to classification models, covering both soft-label and hard-label settings, but also extends to sequence-to-sequence models, such as NMT. It achieves this by providing optimized objectives and customized attack strategies for each model type.
    \item We evaluate our attack method on real-world public datasets, demonstrating that it achieves superior performance in terms of attack effectiveness, imperceptibility, and fluency of examples.
\end{itemize}

The rest of this paper is structured as follows. We review adversarial attacks for NLP models and the applications of Cross-Entropy optimization in Section \ref{related work}. We detail our proposed method in Section \ref{CEA} and then evaluate the performance of the proposed method through empirical analysis in Section \ref{experiments}. We conclude the paper with suggestions for future work in Section \ref{conclusion}.

\section{Related Work} \label{related work}
In this section, we review literature related to this research. 

\subsection{Black-box Adversarial Attacks to NLP models}

Text adversarial attacks have gained attention with the rise of DNNs like BERT \cite{Devlin2019BERTPO}, revealing vulnerabilities in both classifiers and sequence-to-sequence models like NMT. Black-box attacks are divided into soft-label, using predicted labels and confidence scores, and the more challenging hard-label, which relies only on predicted labels.



{Soft-label} attacks use the predicted label and confidence score to generate adversarial examples. Attackers leverage the confidence score to calculate word importance \cite{yang2021bigram, li2021clare, li2020bertattack, ren2019pwws} and apply a greedy mechanism to create adversarial examples. For instance, PWWS \cite{ren2019pwws} ranks word importance and replaces words sequentially until adversarial examples are produced. Other methods employ various optimization algorithms \cite{alzantot2018generating, jia2019faga, zang2020word}. In soft-label settings, predicted labels help determine when adversarial examples are generated. {Hard-label} attacks are more challenging as they only use the predicted label without access to confidence scores. This setting has received less attention, with \citeauthor{hardfirst} being a notable early effort, using a genetic algorithm in a sparse domain to generate high-quality adversarial examples. Texthoaxer \cite{ye2022texthoaxer} addresses budgeted hard-label attacks as a gradient-based optimization problem in the continuous word embedding space. Subsequent works have improved efficiency with simplified word importance ranking \cite{liu2023sspattack, zhu2024limeattack}.

Beyond classifiers, {sequence-to-sequence} NLP models like NMTs, summarization, and question-answering systems, which are used commercially, also face security and fairness concerns. Belinkov \citep{belinkov2017synthetic} initiated attacks on NMT by exploiting natural typos in character-based neural machine translation without assuming any gradients. Tan \cite{Tan2020ItsMT} successfully attacked NMTs and summarization models by using BLEU \cite{bleu} to identify and replace keywords with synonyms, impacting BLEU scores. Seq2Sick \cite{cheng2020seq2sick} achieved similar results by employing a projected gradient method combined with group lasso and gradient regularization, incorporating novel loss functions for non-overlapping and targeted keyword attacks.

\subsection{Optimization for Adversarial Attacks}
Optimization methods for generating textual adversarial attacks can be broadly categorized into ranking-based, genetic-based, and sampling-based approaches, each tailored to address different optimization objectives for adversarial generation.

Ranking-based optimization methods prioritize identifying the most impactful words or phrases for modification, often leveraging gradient-based techniques. These approaches aim to maximize the model's misclassification probability with minimal changes to the input, ensuring semantic coherence. For instance, TextFooler \cite{jin2019textfooler} and DeepWordBug \cite{gao2018blackbox} utilize gradients to rank words by their susceptibility to perturbation. By systematically selecting and replacing words based on their sensitivity scores, these methods effectively alter model predictions while preserving input semantics.

Genetic-based optimization methods, on the other hand, adopt evolutionary strategies to iteratively refine adversarial candidates. Alzantot et al.'s Genetic Algorithm (GA) \cite{alzantot2018generating} for text attacks creates a population of adversarial examples by modifying words or phrases, selecting candidates with the highest misclassification potential for subsequent iterations. This process mimics natural selection, retaining effective mutations while discarding less impactful ones. Particle Swarm Optimization (PSO) \cite{zang2020word} further formulates adversarial substitutions as a combinatorial optimization problem, efficiently searching for perturbations that achieve attack objectives with minimal semantic loss.

Sampling-based optimization methods have shown significant promise in generating effective adversarial attacks. Notable approaches include Metropolis-Hastings Attacks (MHA) \cite{mha}, Fraud's Bargain Attack (FBA) \cite{fba}, and Reversible Jump Attacks (RJA) \cite{rja}. MHA leverages the Metropolis-Hastings algorithm to balance fluency, attack success, and computational efficiency by iteratively refining candidates through insertion, deletion, and replacement operations. FBA enhances this by introducing a Word Manipulation Process (WMP) that stochastically selects positions for perturbation, creating a larger search domain and improving semantic preservation. RJA further advances sampling-based attacks by dynamically adjusting the number of perturbed words through the Reversible Jump algorithm, while incorporating Metropolis-Hastings-based Modification Reduction (MMR) to restore unnecessary changes for improved imperceptibility. These methods effectively navigate the high-dimensional space of adversarial text, providing efficient, adaptive frameworks for crafting high-quality adversarial examples.

Overall, the choice of optimization method depends on the specific requirements of the attack, with sampling-based methods showing promise for their adaptability and efficiency in high-dimensional adversarial settings.

\subsection{Cross-Entropy Optimization}

The Cross-Entropy (CE) optimization method, initially introduced by Rubinstein \cite{rubinstein1999cross}, was developed for rare-event simulation, particularly to estimate probabilities of low-occurrence events. Over time, it has been adapted for optimization tasks, gaining widespread applications in fields such as combinatorial optimization and machine learning. In combinatorial optimization, CE has been used for complex problems like the Traveling Salesman Problem (TSP) \cite{ce_tsp}, Max-Cut \cite{ce_maxcut}, and Quadratic Assignment Problem (QAP) \cite{ce_qap}. These applications leverage the CE method's probabilistic sampling framework to refine solutions iteratively. In machine learning, CE has been applied to hyperparameter tuning \cite{ce_hyperparam}, feature selection \cite{ce_feature}, and neural architecture search (NAS) \cite{ce_nas}, demonstrating its utility in exploring high-dimensional, non-convex spaces efficiently. Its extension to adversarial attacks in NLP and image domains highlights its ability to generate impactful perturbations by treating adversarial generation as a rare-event simulation \cite{ce_adv_1, ce_adv_2}.

Mathematically, the CE optimization method operates on the principle of iteratively refining a distribution to approximate the maximum of a target function. Consider the probability \(\ell = \mathbb{P}(X \geq \gamma)\), where \(X\) is a random variable, and \(\gamma\) defines a rare-event threshold. CE estimates \(\ell\) through iterative sampling and updates based on the following formulation:
\begin{align}
    \text{Estimate } \mathbb{P}(S(X) \geq \gamma),
\end{align}
where \(S(X)\) is a real-valued function. For optimization, this is reformulated as:
\begin{align}
    \text{Determine } \max_{\boldsymbol{x} \in \mathcal{X}} S(\boldsymbol{x}),
\end{align}
where \(\boldsymbol{x}\) represents the decision variable. By iteratively refining the probability distribution toward high-performing regions of the search space, CE effectively identifies rare or optimal solutions. This mathematical foundation makes CE particularly suitable for adversarial attacks, where perturbations are treated as rare events requiring targeted exploration.

The potential of CE for adversarial attack generation is supported by its success in optimization-based methods, widely used in crafting adversarial examples \cite{alzantot2018generating, zang2020word}. While existing approaches like genetic algorithms and particle swarm optimization often struggle with high computational costs or limited adaptability, CE excels through its probabilistic framework, treating adversarial examples as rare events. This aligns naturally with the challenges of adversarial attacks, where subtle perturbations must effectively degrade model performance while preserving semantic integrity. By iteratively sampling, evaluating, and refining candidates, CE efficiently navigates the high-dimensional input space to identify optimal adversarial perturbations, making it a robust and principled method for tackling adversarial attack generation \cite{ce_adv_1, ce_adv_2}.

\section{Cross-Entropy Attacks}\label{CEA}
In this section, we introduce our proposed Cross-Entropy Attacks (CEA) model. Given a textual model $F$, an adversary aims to generate an adversarial document $\mathbf{x'}$ from adversarial candidates $\mathbf{x}$ based on an input $ x=[w_1, \ldots w_i, \ldots w_n]$ with $n$ tokens and true label $y$. Specifically, the attacker tend to optimize the an customised function $f(\mathbf{x}\vert x)$,
\begin{align}
    \mathbf{x'}=\argmax f(\mathbf{x}\vert x)
\end{align}
so that classifiers make wrong predictions, and NMT outputs low-quality translations.

\subsection{Objective Functions}
The general setting for attacking the NLP models is to deprive the performance of the models while reserving the original meaning and modifying limited words. Therefore, we propose the general objective function $f(\mathbf{x}|x)$ for the adversarial candidate $\mathbf{x}$:
\begin{align}
        f(\mathbf{x}|x)=m\left(F(\mathbf{x})\right)\cdot \text{Sem}\left(\mathbf{x}|x\right) \label{eqt: obj}\\
        \nonumber \text{s.t} \quad \text{Sem}\left(\mathbf{x}\vert x\right) \leq \epsilon \quad \text{and} \quad R\left(\mathbf{x}\vert x\right) \leq \eta,
\end{align} where $m(\cdot) \in (0,1)$ is the measure for model performance which will vary for attacking different NLP models, and we will detail in the following sub-sections. $R\left(\mathbf{x}|x\right)$ is the modification rate.


\begin{algorithm}[t!]
\caption{The Cross-Entropy Attack Algorithm}
\label{algo: ce attack}
\begin{algorithmic}[1]
    \State \textbf{Input:} Text $\mathbf{x}$, substitution sets $\mathbf{s}_i$, maximum iterations $T$, number of adversarial candidates $N$
    \State \textbf{Output:} Adversarial example $\mathbf{x'}$
    
    \State \textbf{Initialize} $\hat{\mathbf{p}}_0$ with uniform distribution with probability $\frac{1}{n_i}$, $\hat{\gamma}_0$ with a fixed value $0.5$
    
    \For{$t = 0$ to $T$}
        \For{$h = 1$ to $N$}
            \State \textbf{Sampling}: Create $\mathbf{x}^t_h$ by sampling from $\mathrm{Cat}(\Theta)$ with probability $p(\mathbf{x} \mid \Theta)$
            \State Calculate objective $f(\mathbf{x}^t_h \mid \mathbf{x})$ using Eq \ref{eqt: obj} with performance measures: 
            \Statex \quad - Eq \ref{eqt: soft performance} for soft-label classifiers
            \Statex \quad - Eq \ref{eqt: hard performance} for hard-label classifiers
            \Statex \quad - Eq \ref{eqt: NMT measure} for NMTs
        \EndFor
        
        \State Order the performances: $f_{(1)} \leq \ldots \leq f_{(N)}$
        \State \textbf{Update} $\hat{\gamma}^{t+1}$ to the $(1 - \rho)$-quantile of the performances using Eq \ref{eqt: update rho}
        \State \textbf{Update} $\hat{p}_{(i, j)}^{t+1}$ using Eq \ref{eqt: update p}
    \EndFor
    
    \For{each word $w_i$}
        \State $s'_i = \argmax_{s_{(i,j)} \in S_i } \hat{p}^T_{(i,j)}$
    \EndFor
    
    \State Construct the adversarial example $\mathbf{x'} = (s'_1, \ldots, s'_n)$
\end{algorithmic}
\end{algorithm}

\subsubsection{Objective Function for Classifiers}
We provide objective functions for both soft-label and hard-label black-box setting by constructing the performance measure, $m(\cdot)$. Supposing that the victim $F$ is with $K$ classes and $F_{c} \in [0,1]$ is the confidence of predicting the correct class, we customize the measure of the attacking performance for both soft-label and hard-label in Eq \ref{eqt: soft performance} and \ref{eqt: hard performance}:
\begin{align}
     \text{soft-label:}\quad m &\left(F(\mathbf{x})\right) = 
    \begin{cases}
        \displaystyle 1 - F_c(\mathbf{x})  > \frac{1}{K} \\[0.8em]
        \displaystyle 1 - \frac{1}{K}   \leq \frac{1}{K}
    \end{cases}, \label{eqt: soft performance} \\ 
    \text{hard-label:}\quad m &\left(F(\mathbf{x})\right) = \mathbf{1}(F(\mathbf{x})\neq y) \label{eqt: hard performance}
\end{align}
We argue that the goal of attacking the classifier is to flip the label rather than extensively decrease confidence. Over-decreasing the models' confidence will come with a sacrifice of imperceptibility and quality. Eq. \ref{eqt: soft performance} specifies a lower truncated function at value $\frac{1}{K}$, which means the attacker will get no bonus reward for steadily bullying the victims. The cut-off value $\frac{1}{K}$ is chosen based on the pigeon-hole principle \cite{pigeonhole}, ensuring a consistent $ m(F(\mathbf{x})) \leq \frac{1}{K}$ value will guarantee a misclassified examples.

\subsubsection{Objective Function for NMT}
Unlike textual classifiers, adversarial attacks on sequence-to-sequence models by altering salient words in the input text to largely deprave the output, as opposed to simply flipping the labels as with classifiers \cite{morphin2020tan}. To this end, we propose a different performance measure in Eq \ref{eqt: NMT measure}:

\begin{align}
    m\left(F(\mathbf{x})\right) &= 1 - \text{BLEU}\left(F(\mathbf{x}) \vert y\right) \cdot \text{Sem}\left(F(\mathbf{x}) \vert y\right) \label{eqt: NMT measure}
\end{align}
The function \( m(F(x^{\prime})) \) measures the degradation of NMT by combining two components: the BLEU score and semantic similarity, targeting different aspects of NMT. The BLEU score \cite{bleu} based on n-gram precision, assesses word alignments while $\text{Sem}(F(\mathbf{x}) \vert y) $ represents the semantic similarity between the generated and target translations. By including both terms, we ensure that the degraded translation maintains semantic coherence while attacking word alignment and semantics.

\subsection{Sememe-associated Word Candidates}
To perform sememe-level attacks, we adopt a framework leveraging the granularity of sememes as defined by HowNet \cite{dong2010hownet,qi2019openhownet}. Sememes, the smallest semantic units, enable precise and meaningful perturbations that align with the contextual and syntactic structure of the input. For each sememe-associated word \( w_i \), we construct substitution candidate sets \(\mathbf{s}_i = \{s_{(i,1)}, \ldots, s_{(i,n_i)}\}\) by integrating context-aware suggestions from Masked Language Models (MLMs) and sememe-guided thesauri. First, the input text is segmented into its sememe-level components, and \( w_i \) is masked to create \(\mathbf{x}_{\text{mask}} = x \backslash w_i = (w_1, \ldots, w_{i-1}, w_{i+1}, \ldots, w_n)\). The masked input is processed through an MLM \(\mathcal{M}\) to generate the top \( K \) candidates \(\mathbf{s}_i^{\mathcal{M}}\). Concurrently, we derive a synonym set \(\mathbf{s}_i^{\text{syn}}\) from HowNet-based thesauri, which utilize sememe definitions to ensure semantic consistency. The final substitution set is obtained as the intersection \(\mathbf{s}_i = \mathbf{s}_i^{\mathcal{M}} \cap \mathbf{s}_i^{\text{syn}}\), ensuring that adversarial examples preserve semantic coherence while introducing subtle yet effective perturbations. By operating at the sememe level, this approach enhances the contextual relevance and fluency of adversarial attacks.



\subsection{CE Optimization}
For CE optimization, we first assign a Categorical distribution to each candidate $\mathbf{s_i}$ with the probability mass function (pmf) $p(s_i\vert \theta_i)$, which further implies that the adversarial candidates will follow the multivariate Categorical distribution with pmf $p(\mathbf{x}\vert \Theta)$ as the adversarial candidate $\mathbf{x}$ is constructed by the combination of substitutions $s_i$. The distributions are mathematically formulated in the Eq \ref{eqt: distribution} below:
\begin{align}
    \mathbf{s}_i\sim &\mathrm{Cat}(\theta_i),\quad \mathbf{x} \sim \mathrm{Cat}(\Theta), \label{eqt: distribution}
\end{align}
where
\begin{align*}
 \mathbf{x}=&(s_1,\ldots, s_i, \ldots s_n), \quad s_i \in \mathbf{s}_i,\\
 \Theta=&\{\theta_1,\ldots  \theta_i, \ldots, \theta_n\},\\
 \theta_i =&(p_{(i,1)},\ldots, p_{(i,j)} , \ldots p_{(i, n_i)}), \\
 p_{(i,j)}& \in (0,1),\text{  and  }\sum_{j=1}^{n_i} p_{(i,j)}=1.
\end{align*} $n_i$ is the number of substations of word $w_i$ and $p_{(i,j)}$ denotes the probability for drawing the $j$th substitution $s_{i,j}$ for word $w_i$.

The CE method is based on a sequence of rare event simulations shown in Eq \ref{eqt: CE rare event}:
\begin{align}
\mathbb{P}(f(\mathbf{x}\vert x) \geq \gamma), \label{eqt: CE rare event}
\end{align} where the $\gamma$ is a reasonably large number based on the bounds of the objective function, $0 \leq f(\mathbf{x}\vert x) \leq 1$. The Cross-Entropy (CE) optimization firstly involves creating a sequence of parameters \(\hat{\Theta}_0, \hat{\Theta}_1, \ldots\) and levels \(\hat{\gamma}_1, \hat{\gamma}_2, \ldots\), such that \(\hat{\gamma}_1, \hat{\gamma}_2, \ldots\) converge to the optimal performance (1 here) and \(\hat{\Theta}_1, \hat{\Theta}_2, \ldots\) converge to a constant distributions on the optimal substitutes, which will finally construct the optimal adversarial example $\mathbf{x'}$.

To be more specific, CE attack has three steps: \textbf{initialization}, \textbf{sampling} and \textbf{updating}. We will detail our algorithm step by step. We \textbf{initially} assign a discrete uniform distribution to each substitution set $\mathbf{s}_i$, with $\hat{p}^0_{(i,j)}=\frac{1}{n_i}$, and set $\hat{\gamma}^0=0.5$. The setting $\hat{p}^0_{(i,j)}=\frac{1}{n_i}$ is based on our prior assumption that each substitute has an equal probability to fool the classifier, which is different the word importance rank from the literature \cite{yang2021bigram, li2021clare, zhu2024limeattack}. The value $\hat{\gamma}^0=0.5$ represents the initial threshold for defining a rare event, serving as a neutral starting point. In the iterative process, we start by \textbf{sampling} adversarial candidates. At iteration $t$, we generate $N$ adversarial candidates $\mathbf{x}^t_1, \ldots, \mathbf{x}^t_N$ from the Categorical distributions $\mathrm{Cate}(\hat{\Theta}^t)$. We then evaluate the performance of each candidate with objective function in Eq \ref{eqt: obj}, resulting in the corresponding values $f^t_1, \ldots, f^t_N$. After evaluating the performance of the candidates, we \textbf{update} the parameters for the next iteration. We sort t`  e performance values in ascending order, $f_{(1)} \leq \ldots \leq f_{(N)}$. The new threshold $\hat{\gamma}^{t+1}$ is set to the $(1 - \rho)$-quantile of these values: 
\begin{align}
    \hat{\gamma}^{t+1} = f_{\lceil (1-\rho)N \rceil}. \label{eqt: update rho}
\end{align}
With $\hat{\gamma}^t$ updated, we then revise $\hat{\Theta}^t$ by recalculating its components $\hat{p}^t_{(i,j)}$ using the following equation:

\begin{align}
    \hat{p}^{t+1}_{(i,j)} = \frac{\sum_{h=1}^{N} \mathbf{1}\{f(\mathbf{x}^t_h \vert x) \geq \hat{\gamma}^t\} \mathbf{1}\{s^t_{(i,h)}=s_{(i,j)}\}}{\sum_{h=1}^{N} \mathbf{1}\{f(\mathbf{x}^t_h \vert x) \geq \hat{\gamma}^t\}}, \label{eqt: update p}
\end{align} which estimates $\hat{p}^{t+1}_{(i,j)}$ by calculating the frequency ratio of substitution candidate $s_{(i,j)}$ among the adversarial candidates over the threshold $\hat{\gamma}^t$.

After iteratively running this process for $T$ iterations, we obtain the final output $\hat{\Theta}^T=\{\hat{\theta}^T_1, \ldots, \hat{\theta}^T_{n_i}\}$. By selecting the highest corresponding substitution, we determine the final chosen substitutions and adversarial example, $\mathbf{x'}=(s'_1, \ldots, s'_n)$. The complete algorithm for CEA is provided in Algorithm \ref{algo: ce attack}.

\section{Experiments and Analysis}\label{experiments}
In this section, we comprehensively evaluate the performance of our method against the current state of the art. Besides the main results (Sec. \ref{main results}) of attacking performance and imperceptibility, we also conduct experiments on attacking LLM (Sec. \ref{attack llm}), ablation studies (Sec. \ref{ablation}), performance front of defense mechanism (Sec. \ref{defense}), transferability (Sec. \ref{transfer}), target attacks (Sec. \ref{target}), adversarial retraining (Sec. \ref{retraining}), efficiency analysis (Sec. \ref{efficiency}).

To ensure reproducibility, we provide the code and data used in our experiments in a GitHub repository\footnote{ \url{https://github.com/MingzeLucasNi/CE-Attack.git}}.

\subsection{Experimental Settings}

In our Cross-Entropy Attack (CEA) setup, we employed specific hyperparameters to control the optimization process across all tested NLP tasks. Each iteration of CEA samples 100 adversarial candidates per input, chosen from substitution sets constructed using Masked Language Models (MLMs) combined with a thesaurus-based filter. We set a threshold update rate of \( \rho = 0.5 \), which incrementally adjusts the performance threshold \( \gamma \) at each iteration, ensuring a balance between quality and convergence efficiency. The optimization was capped at a maximum of 50 iterations, based on preliminary experiments indicating that 50 iterations were sufficient for convergence without excessive computational demand. We initialized each substitution candidate with a uniform probability \( \hat{p}_{(i,j)} = \frac{1}{n_i} \), avoiding any prior bias in word replacement likelihood. This initialization allowed the CEA algorithm to effectively explore substitution options from the beginning, yielding robust adversarial examples that adapt to various tasks and model architectures.
\subsection{Datasets}
In this section, we provide detailed information about the datasets used in our experiments.

\subsubsection{Classification}
We evaluate the effectiveness of our methods on four widely-used and publicly available benchmark datasets: AG's News \citep{ag_news}, Emotion \citep{emotion}, SST2 \citep{sst2}, and IMDB \citep{IMDB}. AG's News is a news classification dataset containing 127,600 samples across four topic categories: \textit{World, Sports, Business, Sci/Tech}. SST2 \citep{sst2} is a binary sentiment classification dataset with 9,613 samples labeled as either \textit{positive} or \textit{negative}. The IMDB dataset \citep{IMDB} consists of movie reviews from the Internet Movie Database and is commonly used for binary sentiment classification, categorizing reviews into `positive' or `negative' sentiments. Detailed statistics for these datasets are provided in Table \ref{tab: datasets and models}.

\subsubsection{Translation}
For evaluating attacks on Neural Machine Translation (NMT), we use three publicly available datasets: WMT T1 \citep{TiedemannThottingal:EAMT2020}, WMT T2 \citep{TiedemannThottingal:EAMT2020}, and WMT18 \citep{WMT18}. The WMT T1 and WMT T2 datasets are subsets of the WMT benchmark, specifically focusing on English-to-Chinese (en-zh) translation tasks. These datasets are critical for assessing the robustness of NMT models against adversarial attacks. Both datasets contain English-to-Chinese sentence pairs, allowing for a thorough evaluation of how NMT models handle translation challenges under adversarial conditions. WMT T1 and T2 cover various linguistic structures and vocabulary, making them suitable for testing the adaptability of NMT models to adversarial perturbations. WMT18 further complements these datasets by offering a broader range of translation scenarios, which is essential for validating the generalizability of the attack strategies across different contexts. Detailed statistics for these datasets can be found in Table \ref{tab: datasets and models}.

\subsection{Victim Models}
 We apply our attack algorithm to textual classifiers and NMTs, the details for training these victim models can be found below.
\subsubsection{Victim Classifiers}
The details of the victim model are detailed in this subsection.
\paragraph*{BERT-based Classifiers} We choose three well-performed and popular BERT-based models, which we call BERT-C models (where the letter ``C'' represents ``classifier''),  pre-trained by Huggingface\footnote{\url{https://huggingface.co/}}. Due to the different sizes of the datasets, the structures of BERT-based classifiers are adjusted accordingly. The BERT classifier for AG's News is structured by the \textit{Distil-RoBERTa-base} \cite{Sanh2019DistilBERTAD} connected with two fully connected layers, and it is trained for 10 epochs with a learning rate of 0.0001. For the Emotion dataset, its BERT-C adopts another version of BERT, \textit{Distil-BERT-base-uncased} \cite{Sanh2019DistilBERTAD}, and the training hyper-parameters remain the same as BERT-C for AG's News. Since the SST2 dataset is relatively small compared with the other two models, the corresponding BERT classifier utilizes a small-size version of BERT, \textit{BERT-base-uncased} \cite{Devlin2019BERTPO}. The test accuracy of these BERT-based classifiers before they are under attacks are listed in Table \ref{tab: datasets and models} and these models are publicly accessible\footnote{\url{https://huggingface.co/mrm8488/distilroberta-finetuned-age_news-classification}} \footnote{\url{https://huggingface.co/echarlaix/bert-base-uncased-sst2-acc91.1-d37-hybrid}} \footnote{\url{https://huggingface.co/lvwerra/distilbert-imdb}}.
\par
\paragraph*{TextCNN-based models}The other type of victim model is TextCNN \cite{Kim2014ConvolutionalNN}, structured with a 100-dimension embedding layer followed by a 128-unit long short-term memory layer. This classifier is trained 10 epochs by ADAM optimizer with parameters: learning rate $lr=0.005$, the two coefficients used for computing running averages of gradient and its square are set to be 0.9 and 0.999 $(\beta_1=0.9$, $\beta_2=0.999)$,  the denominator to improve numerical stability $\sigma=10^{-5}$. The accuracy of these TextCNN-base models is also shown in Table \ref{tab: datasets and models}.

\begin{table}[t!]
\small
\caption{Datasets and Performance of victim models. The metrics for classifiers and NMTs are accuracy and BLEU, respectively.}
\centering
\begin{tabular}{ccccccc}
\toprule
Dataset &Size &Avg.Length & Type & Task    &  Model &  Metrics\\
\midrule
\multirow{2}[0]{*}{AG's News}  & \multirow{2}[0]{*}{12,700} & \multirow{2}[0]{*}{37.84}   &\multirow{2}[0]{*}{classifier} & \multirow{2}[0]{*}{News topics}
 & BERT &  94\%    \\
~&~&~     &~&~&  TextCNN   &~ 90\%  \\
\midrule

\multirow{2}[0]{*}{SST2} & 
\multirow{2}[0]{*}{9,613}  &
\multirow{2}[0]{*}{19.31}&
\multirow{2}[0]{*}{classifier}&
\multirow{2}[0]{*}{\shortstack{Sentiment\\analysis}}&
BERT-C      &   91\%   \\
~&~&~&~&~ &  TextCNN   &  83\% \\

\midrule
\multirow{2}[0]{*}{IMDB} & 
\multirow{2}[0]{*}{50,000}  &
\multirow{2}[0]{*}{279.48} & 
\multirow{2}[0]{*}{classifier}&
\multirow{2}[0]{*}{Movie review}
&BERT-C      &   93\%   \\
~&~&~&~&~ &  TextCNN   &  88\% \\

\midrule
\multirow{2}[0]{*}{WMT T1} & 
\multirow{2}[0]{*}{9,000}  &
\multirow{2}[0]{*}{14.10}  &
\multirow{2}[0]{*}{NMT}&
\multirow{2}[0]{*}{\shortstack{Wikipedia \\ \& Reddit}}
&T5      &   36.8   \\
~&~&~&~&~ &  mBART   &  31.4 \\

\midrule
\multirow{2}[0]{*}{WMT T2} & 
\multirow{2}[0]{*}{9,000}  &
\multirow{2}[0]{*}{16.51}  &
\multirow{2}[0]{*}{NMT}&
\multirow{2}[0]{*}{Wikipedia}
&T5      &   31.1   \\
~&~&~&~&~ &  mBART   &  27.1 \\
\bottomrule
\end{tabular}
\label{tab: datasets and models}
\end{table}

\subsubsection{Victim Neural Machine Translations}
To evaluate our attack algorithm on neural machine translation (NMT) models, we assigned English-to-Chinese (simplified) translation tasks to two widely-used and publicly available models: T5 \citep{t5} and mBART \citep{mbart}, both pre-trained on Huggingface\footnote{\url{https://huggingface.co/}}.

\textbf{T5 (Text-To-Text Transfer Transformer)} \citep{t5} is a versatile sequence-to-sequence model that approaches every language processing problem as a text-to-text task. It is trained on a large, diverse corpus and can handle multiple tasks, including translation, summarization, and text generation. In this study, we focus on its application to English-to-Chinese (en-zh) translation tasks. T5's architecture leverages transfer learning, making it highly effective in adapting to various NLP tasks. The T5 models used for WMT T1 and T2 are pre-trained and publicly available on Huggingface\footnote{\url{https://huggingface.co/utrobinmv/t5_translate_en_ru_zh_large_1024}} \footnote{\url{https://huggingface.co/utrobinmv/t5_translate_en_ru_zh_large_1024_v2}}.

\textbf{mBART (Multilingual BART)} \citep{mbart} is another powerful sequence-to-sequence model, designed specifically for multilingual tasks, including translation. Pre-trained on large-scale multilingual corpora, mBART is fine-tuned to generate high-quality translations across various languages. In this study, we use mBART for English-to-Chinese (en-zh) translation, where its robust architecture and multilingual pre-training make it an ideal candidate for evaluating the impact of adversarial attacks. mBART's ability to handle diverse linguistic structures and proficiency in cross-lingual tasks provide a strong platform for testing adversarial robustness in multilingual settings. The mBART models used for WMT T1 and T2 are also pre-trained and publicly available on Huggingface\footnote{\url{https://huggingface.co/Helsinki-NLP/opus-mt-en-zh}} \footnote{\url{https://huggingface.co/liam168/trans-opus-mt-en-zh}}.

Detailed statistics for these NMT models and their respective datasets can be found in Table \ref{tab: datasets and models}.

\subsection{Baselines}
We compare our method against several established baselines across different attack settings:

\paragraph{For classifiers (Soft-label):}
For soft-label attacks, we compare against multiple strong baselines. PSO\citep{zang2020word} utilizes particle swarm optimization to select word candidates from HowNet, iteratively refining substitutions to find optimal adversarial examples. TextFooler \citep{Jin2020IsBR} ranks words by importance and replaces them with semantically similar alternatives until the classification is altered. CLARE~\citep{li2021clare} employs contextual perturbations using masked language models to ensure fluency and semantic consistency. Additionally, we employ the Reversible Jump Attack \citep{rja}, which utilizes an MCMC-based Reversible Jump sampler for generating adversarial examples, enabling comprehensive performance comparison.

\paragraph{For classifiers (Hard-label):}
In hard-label scenarios, we evaluate against three state-of-the-art methods. HLBB \citep{hardfirst} is a boundary-based attack that identifies word importance through gradient estimation. TextHoaxer \citep{ye2022texthoaxer} focuses on efficient word importance ranking in hard-label settings. LimeAttack \citep{zhu2024limeattack} employs structured search to identify critical words for perturbation while maintaining semantic coherence.

\paragraph{For NMTs:}
For translation attacks, we compare the proposed method with several advanced techniques. HAA \citep{haa} combines translation and self-attention mechanisms to identify vulnerable words. Seq2Sick \citep{cheng2020seq2sick} generates adversarial examples while maintaining semantic similarity. Morpheus \citep{Tan2020ItsMT} targets inflectional forms of words to degrade BLEU scores while preserving meaning.

This comparison covers diverse attack strategies, allowing a comprehensive evaluation of our method's effectiveness across different scenarios.

\subsection{Metrics}
To measure these facets, we evaluate from three perspectives: attacking performance, imperceptibility, and example quality.

\subsubsection*{For attacking performance:}
\begin{itemize}
    \item \textbf{Successful Attack Rate (SAR)}: This metric is used for classifiers and is defined as the percentage of successful adversarial examples that cause the model to make an incorrect prediction. The formulation for SAR is:
    \[
    \text{SAR} = \frac{\text{Number of successful adversarial examples}}{\text{Total number of adversarial examples}}
    \]
    \item \textbf{BLEU Drop (BD)}: Applied to NMTs, BLEU Drop quantifies the reduction in the BLEU score due to the adversarial attack. It is calculated as the difference between the BLEU score of the original output and the BLEU score of the adversarial output:
    \[
    \text{BD} = \text{BLEU}_{\text{original}} - \text{BLEU}_{\text{adversarial}}
    \]
    where \(\text{BLEU}_{\text{original}}\) is the BLEU score of the model's original translation and \(\text{BLEU}_{\text{adversarial}}\) is the BLEU score after the attack.
    \item \textbf{Semantic Drop (SD)}: Also used for NMTs, Semantic Drop measures the decrease in semantic similarity between the original translation and the adversarial translation. It is calculated using the cosine similarity between sentence embeddings:
    \[
    \text{SD} = \text{Sim}_{\text{original}} - \text{Sim}_{\text{adversarial}}
    \]
    where \(\text{Sim}_{\text{original}}\) is the cosine similarity of the original sentence embedding and \(\text{Sim}_{\text{adversarial}}\) is the cosine similarity of the adversarial sentence embedding.
\end{itemize}

\subsubsection*{For imperceptibility:}
\begin{itemize}
    \item \textbf{Modification Rate (Mod)}: This metric calculates the percentage of tokens that have been modified in the adversarial example:
    \[
    \text{Mod} = \frac{\text{Number of modified tokens}}{\text{Total number of tokens}} 
    \]
    \item \textbf{Semantic Similarity (SS)}: Measured by the Universal Sentence Encoder (USE) \citep{cer2018universal}, this metric assesses the cosine similarity between the embeddings of the original and adversarial sentences, indicating how closely the adversarial example resembles the original text.
\end{itemize}

\subsubsection*{For example quality:}
\begin{itemize}
    \item \textbf{Perplexity (PPL)} \citep{radfordlanguage}: This metric, based on GPT-2, evaluates the fluency of the adversarial examples by calculating how well the language model predicts the sequence of words.
    \item \textbf{Grammar Error Rate (GE)}: Measured using LanguageTool \citep{naber2003rule}, this metric calculates the absolute increase in grammatical errors in successful adversarial examples compared to the original text.
\end{itemize}

\subsection{Experimental Results and Analysis} \label{main results}
\begin{table}[!ht]
    \centering
    \caption{Soft-label attack results: Comparative performance of baselines and CE attack across AG News, SST2 and IMDB datasets. Metrics include Success Attack Rate (SAR), Modification Rate (Mod), Semantic Similarity (SS), Perplexity (PPL), and Grammar Error Rate (GE). Best results are in bold.}
    \small
    \begin{tabular}{p{1.5cm}p{1.5cm}p{1.5cm}ccccc}
    \toprule
        \multirow{2}{*}{Datasets} & \multirow{2}{*}{Victims} & \multirow{2}{*}{Attacks} & \multicolumn{5}{c}{Metric}\\ \cmidrule(lr){4-8}
        ~ & ~ & ~ & SAR & Mod & SS & PPL & GE \\ \midrule
        \multirow{10}{*}{AG News} & \multirow{5}{*}{BERT} & CLARE & 71.3 & 14.6 & 66 & 233 & 0.21   \\ 
        ~ & ~ & TF & 89.3 & 21.6 & 77 & 312 & 0.28   \\ 
        ~ & ~ & PSO & 93.4 & 21.6 & 69 & 292 & 0.31   \\ 
        ~ & ~ & RJA & 95.1 & 11.1 & 77 & 155 & 0.21   \\ 
        ~ & ~ & \textbf{CEA} & \textbf{98.0} & \textbf{11.0} & \textbf{79} & \textbf{140} & \textbf{0.18}   \\ \cmidrule(lr){2-8}
        ~ & \multirow{5}{*}{TextCNN} & CLARE & 79.3 & 14.2 & 71 & 213 & 0.19   \\ 
        ~ & ~ & TF & 77.3 & 19.4 & 74 & 199 & 0.21   \\ 
        ~ & ~ & PSO & 76.2 & 15.5 & 77 & 142 & 0.14   \\ 
        ~ & ~ & RJA & 88.3 & 11.4 & 79 & 154 & 0.17   \\ 
        ~ & ~ & \textbf{CEA} & \textbf{93.8} & \textbf{10.9} & \textbf{82} & \textbf{127} & \textbf{0.11}   \\ \midrule
        \multirow{10}{*}{SST2} & \multirow{5}{*}{BERT} & CLARE & 94.9 & 22.0 & 74 & 255 & 0.19   \\ 
        ~ & ~ & TF & 90.4 & 15.3 & 80 & 191 & 0.17   \\ 
        ~ & ~ & PSO & 96.6 & 17.2 & 81 & 197 & 0.27   \\ 
        ~ & ~ & RJA & 96.4 & 15.3 & 78 & 201 & 0.18   \\ 
        ~ & ~ & \textbf{CEA} & \textbf{96.5} & \textbf{10.1} & \textbf{88} & \textbf{169} & \textbf{0.13}   \\ \cmidrule(lr){2-8}
        ~ & \multirow{5}{*}{TextCNN} & CLARE & 85.6 & 17.1 & 70 & 174 & 0.17   \\  
        ~ & ~ & TF & 91.0 & 17.2 & 75 & 194 & 0.21   \\  
        ~ & ~ & PSO & 92.2 & 17.2 & 81 & 145 & 0.14   \\  
        ~ & ~ & RJA & 94.0 & 17.2 & 77 & 164 & 0.18   \\  
        ~ & ~ & \textbf{CEA} & \textbf{99.3} & \textbf{10.1} & \textbf{84} & \textbf{141} & \textbf{0.14}   \\ \midrule
        \multirow{10}{*}{IMDB} & \multirow{5}{*}{BERT} & CLARE & 95.1 & 8.3 & 84 & 101 & 0.18   \\  
        ~ & ~ & TF & 96.3 & \textbf{5.6} & 81 & 111 & 0.22   \\  
        ~ & ~ & PSO & 99.1 & 6.3 & 80 & 91 & 0.22   \\  
        ~ & ~ & RJA & 92.1 & 5.9 & 83 & 77 & \textbf{0.19}   \\  
        ~ & ~ & \textbf{CEA} & \textbf{100.0} & \textbf{5.6} & \textbf{93} & \textbf{61} & 0.19   \\ \cmidrule(lr){2-8}
        ~ & \multirow{5}{*}{TextCNN} & CLARE & 98.4 & 6.3 & 80 & 89 & 0.20   \\  
        ~ & ~ & TF & 99.1 & 6.9 & 90 & 166 & 0.22   \\  
        ~ & ~ & PSO & 98.5 & 9.3 & 88 & 164 & 0.28   \\  
        ~ & ~ & RJA & 95.0 & 6.6 & 90 & 89 & 0.23   \\  
        ~ & ~ & \textbf{CEA} & \textbf{100.0} & \textbf{6.2} & \textbf{92} & \textbf{68} & \textbf{0.21}   \\ \bottomrule
    \end{tabular}
    \label{tab: soft-label performance}
\end{table}

\begin{table}[!ht]
    \centering
    \caption{Hard-label attack results: Comparative performance of baselines and CE attack across AG News, SST2 and IMDB datasets using metrics SAR, Mod, SS, PPL, and GE. Best results are in bold.}
    \small
    \begin{tabular}{p{1.7cm}p{1.7cm}p{2cm}ccccc}
    \toprule
        \multirow{2}{*}{Datasets} & \multirow{2}{*}{Victims} & \multirow{2}{*}{Attacks} & \multicolumn{5}{c}{Metric}\\ \cmidrule(lr){4-8}
        ~ & ~ & ~ & SAR & Mod & SS & PPL & GE \\ \midrule
        \multirow{10}{*}{AG News} & \multirow{5}{*}{BERT} & TextHoaxer & 32.8 & 9.6 & 89 & 151 & 0.13   \\ 
        ~ & ~ & HLBB & 35.2 & 7.8 & \textbf{91} & 149 & \textbf{0.10}   \\ 
        ~ & ~ & LimeAttack & 36.1 & 9.8 & 88 & 190 & 0.14   \\ 
        ~ & ~ & \textbf{CEA} & \textbf{36.9} & \textbf{7.7} & \textbf{91} & \textbf{126} & \textbf{0.10}   \\ \cmidrule(lr){2-8}
        ~ & \multirow{5}{*}{TextCNN} & TextHoaxer & 35.7 & 10.9 & 93 & 176 & 0.15   \\ 
        ~ & ~ & HLBB & 37.7 & 7.9 & 88 & 241 & 0.20   \\ 
        ~ & ~ & LimeAttack & 32.8 & 8.8 & 90 & 140 & 0.16   \\ 
        ~ & ~ & \textbf{CEA} & \textbf{43.8} & \textbf{5.7} & \textbf{94} & \textbf{111} & \textbf{0.11}   \\ \midrule
        \multirow{10}{*}{SST2} & \multirow{5}{*}{BERT} & TextHoaxer & 43.4 & 10.7 & 86 & 173 & 0.14   \\ 
        ~ & ~ & HLBB & 31.1 & 12.5 & 84 & 165 & 0.16   \\ 
        ~ & ~ & LimeAttack & 35.1 & 11.5 & 87 & 152 & 0.16   \\ 
        ~ & ~ & \textbf{CEA} & \textbf{47.4} & \textbf{10.3} & \textbf{89} & \textbf{121} & \textbf{0.07}   \\ \cmidrule(lr){2-8}
        ~ & \multirow{5}{*}{TextCNN} & TextHoaxer & 58.4 & 9.5 & 81 & 163 & 0.13   \\ 
        ~ & ~ & HLBB & 63.4 & 8.5 & 81 & 181 & 0.18   \\ 
        ~ & ~ & LimeAttack & 60.4 & 10.5 & 88 & 141 & 0.10   \\ 
        ~ & ~ & \textbf{CEA} & \textbf{71.4} & \textbf{7.9} & \textbf{89} & \textbf{129} & \textbf{0.09}   \\ \midrule
        \multirow{10}{*}{IMDB} & \multirow{5}{*}{BERT} & TextHoaxer & 68.1 & 8.6 & 91 & 65 & 0.21   \\ 
        ~ & ~ & HLBB & 68.3 & 5.0 & \textbf{92} & 78 & 0.18   \\ 
        ~ & ~ & LimeAttack & 73.1 & 5.6 & 90 & 72 & 0.19   \\ 
        ~ & ~ & \textbf{CEA} & \textbf{76.0} & \textbf{4.5} & \textbf{92} & \textbf{54} & \textbf{0.12}   \\ \cmidrule(lr){2-8}
        ~ & \multirow{5}{*}{TextCNN} & TextHoaxer & 76.0 & 4.2 & 90 & 72 & 0.16   \\ 
        ~ & ~ & HLBB & 71.8 & 5.9 & 90 & 77 & 0.16   \\ 
        ~ & ~ & LimeAttack & 74.1 & 5.5 & 85 & 83 & 0.21   \\ 
        ~ & ~ & \textbf{CEA} & \textbf{79.9} & \textbf{5.5} & \textbf{91} & \textbf{62} & \textbf{0.09}   \\ \bottomrule
    \end{tabular}
    \label{tab: hard-label performance}
\end{table}

\begin{table}[!ht]
    \centering
    \caption{NMT attack results: Comparative performance of baselines and CE attack across WMT T1 and T2 datasets using metrics BD, SD, Mod, SS, PPL, and GE. Best results are in bold.}
    \small
    \begin{tabular}{p{1.5cm}p{1.5cm}p{1.5cm}cccccc}
    \toprule
        \multirow{2}{*}{Datasets} & \multirow{2}{*}{Victims} & \multirow{2}{*}{Attacks} & \multicolumn{6}{c}{Metric}\\ \cmidrule(lr){4-9}
        ~ & ~ & ~ & BD & SD & Mod & SS & PPL & GE \\ \midrule
        \multirow{8}{*}{WMT T1} & \multirow{4}{*}{mBART} & Seq2sick & 7 & 8 & 18 & 75 & 369 & 0.21   \\ 
        ~ & ~ & Morph & 11 & 16 & 17 & 83 & 268 & 0.20   \\ 
        ~ & ~ & HAA & 14 & 21 & 19 & 83 & 269 & 0.22   \\ 
        ~ & ~ & \textbf{CEA} & \textbf{17} & \textbf{24} & \textbf{15} & \textbf{85} & \textbf{199} & \textbf{0.15}   \\ \cmidrule(lr){2-9}
        ~ & \multirow{4}{*}{T5} & Seq2sick & 7 & 8 & 18 & 77 & 298 & 0.25   \\ 
        ~ & ~ & Morph & 12 & 11 & 15 & 81 & 241 & 0.23   \\ 
        ~ & ~ & HAA & 16 & 20 & 19 & 86 & 210 & 0.15   \\ 
        ~ & ~ & \textbf{CEA} & \textbf{20} & \textbf{25} & \textbf{14} & \textbf{88} & \textbf{188} & \textbf{0.16}   \\ \midrule
        \multirow{8}{*}{WMT T2} & \multirow{4}{*}{mBART} & Seq2sick & 8 & 11 & 19 & 75 & 477 & 0.26   \\ 
        ~ & ~ & Morph & 12 & 16 & 17 & 84 & 498 & 0.27   \\ 
        ~ & ~ & HAA & 16 & 23 & 18 & 83 & 448 & 0.19   \\ 
        ~ & ~ & \textbf{CEA} & \textbf{17} & \textbf{25} & \textbf{16} & \textbf{85} & \textbf{398} & \textbf{0.18}   \\ \cmidrule(lr){2-9}
        ~ & \multirow{4}{*}{T5} & Seq2sick & 10 & 9 & 21 & 72 & 479 & 0.26   \\ 
        ~ & ~ & Morph & 13 & 11 & 17 & 86 & 432 & 0.21   \\ 
        ~ & ~ & HAA & 17 & 30 & \textbf{14} & 83 & 398 & 0.23   \\ 
        ~ & ~ & \textbf{CEA} & \textbf{16} & \textbf{31} & \textbf{14} & \textbf{90} & \textbf{377} & \textbf{0.20}   \\ \bottomrule
    \end{tabular}
    \label{tab: NMT performance}
\end{table}

\begin{table*}[ht]
\centering
\caption{Exhibition of soft-label adversarial examples by attacking BERT on IMDB. The italic and bolded words denote the original words and their substitutions, respectively.}
\begin{tabular}{p{1.5cm}p{9cm}|l}
\toprule
Attacks & Text & Success \\ \midrule
TF & .... Again, \textit{much} like The Element of Crime, the film ends with our hero unable to \textit{wake up} \textbf{escape} from his nightmare state, left in this terrible place, .... & No \\ \midrule
CLARE & .... Again, much like The Element of Crime, the film ends with our hero unable to \textit{wake up} \textbf{escape} from his nightmare state, left in this terrible place, .... & No \\ \midrule
PSO & .... Again, much like The Element of Crime, the film ends with our hero unable to \textit{wake up} \textbf{recover} from his nightmare state, left in this terrible place, .... & No \\ \midrule
RJA & .... Again, much like The Element of \textit{Crime} \textbf{Death}, the film ends with our hero unable to wake up from his nightmare state, left in this terrible place, .... & No \\ \midrule
\textbf{CEA} & .... Again, \textit{much} \textbf{just} like The Element of Crime, the film ends with our hero unable to wake up from his nightmare state, left in this terrible place, .... & \textbf{Yes} \\ \bottomrule
\end{tabular}
\label{tab: soft-label_examples}
\end{table*}

\begin{table}[ht]
\centering
\caption{Exhibition of hard-label Adversarial Examples by attacking BERT on SST.  The italic and bold words denote the original words and substitutions, respectively.}
\begin{tabular}{p{1.4cm}p{8cm}|p{0.8cm}}
\toprule
Attacks & Text & Success \\ \midrule

TextHoaxer & The history is fascinating; the action is \textit{dazzling} \textbf{attractive} .& No \\ \midrule

HLBB & The history is \textit{fascinating} \textbf{attractive}; the action is dazzling. & No \\ \midrule

LimeAttack & The history is fascinating; the action is \textit{dazzling} \textbf{}. & No \\ \midrule

\textbf{CEA} & The history is fascinating; the action is \textit{dazzling} \textbf{magnificent}.& \textbf{Yes}\\ \bottomrule
\end{tabular}
\label{tab: hard-label examples}
\end{table}

\begin{table*}[ht]
\centering
\caption{Exhibition of Adversarial Examples for attacking mBART on WMT T2. The italic and bold words denote the original words and substitutions, respectively.}
\begin{tabular}{p{1.2cm}p{9cm}|p{1cm}}
\toprule
Attacks & Text & Result \\ \midrule
4Seq2sick & General Edward Hand , the local military commander , banished Col. Croghan from the frontier in 1777 on suspicion of \textit{treason} \textbf{corruption} . & BD:7, SD:10 \\ \midrule

Morph & General Edward Hand , the local military commander , banished Col. Croghan from the frontier in 1777 on \textit{suspicion} \textbf{charges} of treason . & BD:11, SD:12 \\ \midrule

HAA & General \textit{Edward} \textbf{william} Hand , the local military commander , banished Col. Croghan from the frontier in 1777 on suspicion of treason . & BD:4, SD:8 \\ \midrule

\textbf{CEA} & \textit{General} \textbf{Admiral} Edward Hand , the local military commander , banished Col. Croghan from the frontier in 1777 on suspicion of treason . & \textbf{BD:19, SD:28 }\\ \bottomrule
\end{tabular}
\label{tab: nmt examples}
\end{table*}

The main experimental results of the soft-label, hard-label attacks and attacks to NMTs are listed in Table \ref{tab: soft-label performance}, \ref{tab: hard-label performance} and \ref{tab: NMT performance}, respectively. Moreover, we demonstrate adversarial examples crafted by various methods for attacking different models shown in Table \ref{tab: soft-label_examples}, \ref{tab: hard-label examples} and \ref{tab: nmt examples}. We will manifest the performance of the proposed methods in terms of three aspects: attacking performance, imperceptibility and example quality.

\subsubsection{Attacking Performance}
Our adaptive CE attack demonstrates superior performance compared to static baselines in terms of attack success, imperceptibility, and sentence quality. As shown in Tables \ref{tab: soft-label performance} and \ref{tab: hard-label performance}, the CE attack consistently achieves higher success attack rates (SAR) across various datasets and models in both soft-label and hard-label settings. Additionally, the CE attack excels in sequence-to-sequence models such as NMT, as highlighted in Table \ref{tab: NMT performance}, where it surpasses static baselines in attack performance. In terms of imperceptibility, the CE attack achieves the lowest Mod and highest SS, requiring fewer changes and high smenatics preservation to achieve successful attacks, making the alterations less noticeable. Regarding sentence quality, the CE attack also maintains good fluency (PPL) and less grammar error (GE) ensuring that the adversarial examples remain more coherent and natural.

\subsubsection{Imperceptibility Analysis}
The experimental results across Tables \ref{tab: soft-label performance}, \ref{tab: hard-label performance}, and \ref{tab: NMT performance} demonstrate the superior imperceptibility of our CE attack. The CE attack consistently achieves higher semantic similarity while requiring fewer word modifications across all scenarios compared to baseline methods.

This superior imperceptibility stems from CE attack's optimization framework. By leveraging the cross-entropy method with sememe-based word substitutions, CE attack effectively balances between attack success and imperceptibility constraints. The algorithm iteratively refines the probability distribution over word substitutions using Eq \ref{eqt: update p}, converging toward optimal candidates that require minimal modifications while preserving semantic meaning. Additionally, the objective function in Eq \ref{eqt: obj} explicitly incorporates semantic similarity constraints, ensuring the generated adversarial examples maintain high fidelity to the original text. The integration of context-aware MLM suggestions with sememe-guided thesauri further helps select substitutions that preserve both local fluency and global semantic coherence.

\subsubsection{Maintaining Linguistic Quality}
The CE attack maintains superior sentence quality in terms of fluency and grammatical correctness, as evidenced by the Perplexity (PPL) and Grammar Error Rate (GE) metrics. For example, in soft-label attacks on AG News with BERT, CE achieves a PPL of 140 and a GE of 0.18, outperforming RJA and PSO. Similarly, in NMT tasks on WMT T1 with T5, CE achieves a PPL of 188 and GE of 0.16, demonstrating its robustness across different tasks.

The high sentence quality is attributed to CE's reliance on language model guidance and adaptive refinement mechanisms, which prevent linguistic degradation during adversarial generation. This ensures that the crafted adversarial examples are not only effective but also maintain a natural and fluent structure, making them more practical for real-world applications.

The CE attack excels in attacking performance, imperceptibility, and sentence quality across various datasets and models. Its adaptive optimization framework enables it to balance high attack success with minimal perturbations and superior linguistic coherence, establishing CE as a state-of-the-art adversarial attack method.

\subsection{Attack to Large Language Models}\label{attack llm}
Large Language Models (LLMs) have demonstrated remarkable capabilities across diverse tasks but remain vulnerable to adversarial attacks. Unlike traditional NLP models, attacking LLMs requires considering multiple task categories including commonsense reasoning, knowledge recall, reading comprehension, mathematical reasoning, and code generation. We evaluate these attacks using task-specific metrics: accuracy for knowledge and reasoning tasks, perplexity (PPL) for generation fluency, and BLEU/ROUGE scores for comprehension tasks. Our approach employs customized Cross-Entropy (CE) optimization objectives for each task type, enabling targeted evaluation of specific LLM vulnerabilities while providing insights into their overall adversarial robustness.

\subsubsection{Attacking LLM for Classification}

\begin{table}[ht]
\caption{Performance comparison of hard-label adversarial attacks on LLM classification tasks using AG News dataset. Best results in bold.}
\centering
\begin{tabular}{lcccccc}
\toprule
Model & Method & SAR↑ & Mod↓ & SS↑ & PPL↓ & GE↓ \\
\midrule
\multirow{3}{*}{Llama-3.2} & LimeAttack & 32.1 & 11.8 & 0.79 & 168 & 0.21 \\
 & HLBB & 34.5 & 10.2 & 0.81 & 155 & 0.18 \\
 & \textbf{CEA} & \textbf{37.8} & \textbf{8.9} & \textbf{0.84} & \textbf{142} & \textbf{0.15} \\
\midrule
\multirow{3}{*}{Gemma-2} & LimeAttack & 30.6 & 12.4 & 0.78 & 172 & 0.23 \\
 & HLBB & 33.2 & 10.8 & 0.80 & 161 & 0.19 \\
 & \textbf{CEA} & \textbf{36.5} & \textbf{9.3} & \textbf{0.83} & \textbf{148} & \textbf{0.16} \\
\bottomrule
\end{tabular}
\label{tab: llm_classification}
\end{table}

We evaluate our hard-label attack against two popular open-source LLMs: Llama-3.2 (3B parameters) \citep{2024llama} and Gemma-2 (2B parameters) \citep{gemma} on the AG News dataset. These models represent recent advancements in efficient, lightweight LLMs suitable for various downstream tasks.

As shown in Table \ref{tab: llm_classification}, CE attack demonstrates consistent superiority over baseline methods across all evaluation metrics for both LLMs. The improved performance can be attributed to two key factors. First, CE attack's sememe-based word substitution strategy is particularly effective against LLMs' contextualized understanding, enabling more precise and semantically coherent perturbations. Second, the cross-entropy optimization framework adaptively learns the optimal substitution distributions specific to each LLM's decision boundaries, resulting in more efficient attacks with fewer modifications while maintaining better text quality and semantic preservation.

\subsubsection{Attacking LLM for NMT}
We evaluate our approach on the WMT T2 dataset against the same LLMs from the classification victim models, testing their English-to-Chinese translation capabilities. Results are presented in Table \ref{tab: llm_nmt}.

\begin{table}[ht]
\caption{Performance comparison of adversarial attacks on LLM translation tasks using WMT T2 dataset. Best results in bold.}
\centering
\begin{tabular}{lccccccc}
\toprule
Model & Method & BD↑ & SD↑ & Mod↓ & SS↑ & PPL↓ & GE↓ \\
\midrule
\multirow{3}{*}{Llama-3.2} & Seq2Sick & 11.2 & 14.8 & 19.5 & 0.74 & 289 & 0.25 \\
 & MorphAttack & 13.6 & 16.9 & 17.8 & 0.77 & 245 & 0.21 \\
 & \textbf{CEA} & \textbf{15.8} & \textbf{18.7} & \textbf{15.6} & \textbf{0.81} & \textbf{212} & \textbf{0.17} \\
\midrule
\multirow{3}{*}{Gemma-2} & Seq2Sick & 10.5 & 13.9 & 20.1 & 0.73 & 298 & 0.26 \\
 & MorphAttack & 12.8 & 15.8 & 18.4 & 0.75 & 256 & 0.23 \\
 & \textbf{CEA} & \textbf{14.9} & \textbf{17.6} & \textbf{16.2} & \textbf{0.79} & \textbf{228} & \textbf{0.19} \\
\bottomrule
\end{tabular}
\label{tab: llm_nmt}
\end{table}

Results on the WMT T2 dataset show that our CE attack achieves higher BLEU Drop (BD) and Semantic Drop (SD) while requiring fewer modifications. The improved semantic similarity scores and lower perplexity and grammar error rates indicate that our method maintains better meaning preservation and text quality while successfully degrading translation performance. This demonstrates the effectiveness of our approach in generating adversarial examples that can significantly impact LLM translation performance while maintaining text naturalness and grammatical correctness.

\subsection{Ablation Studies}\label{ablation}
To further investigate the effectiveness and adaptability of our Cross-Entropy Attack (CEA) method, we conduct comprehensive ablation studies examining both the core components of CEA and its key hyperparameters. We evaluate the effectiveness of our proposed objective functions and optimization approach, comparing them to alternative methods. Additionally, we analyze the impact of various hyperparameters on the attack’s success across classification and neural machine translation (NMT) tasks. These studies provide deeper insights into how each design choice contributes to overall performance, guiding future optimization.
·

\subsubsection{Effectiveness of CE Optimization} 
\begin{figure}[t!]
    \centering
    \includegraphics[width=0.9\columnwidth]{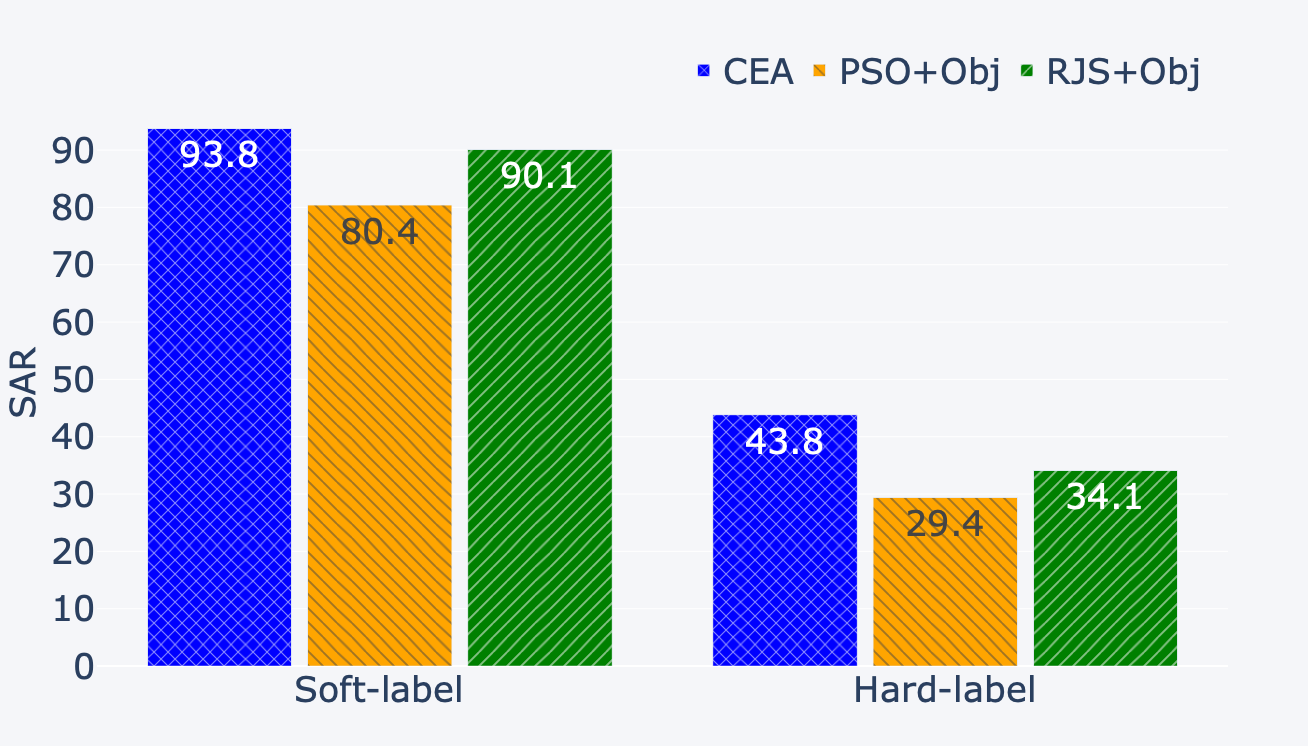}
    \caption{SAR comparison of CEA, PSO+Obj, and RJS+Obj in Soft-label and Hard-label attacks on the AG News dataset using a TextCNN model.}
    \label{fig: optimization}
\end{figure}

To address this question, we compared the attacking performance of the proposed Cross-Entropy (CE) optimization method with Particle Swarm Optimization (PSO) and the Reversible Jump Sampler (RJS) for optimizing the proposed objective functions using a TextCNN model trained on the AG News dataset. The experimental results, presented in Fig \ref{fig: optimization}, show that CE optimization outperformed both PSO and RJS. 

We believe PSO may struggle with sparse spaces, while RJS encounters difficulties with truncated distributions, failing to make Markovian jumps to the next static point. The proposed CE optimization effectively addresses these challenges by minimizing the influence of preset hyper-parameters, narrowing the search space to substitution sets, and enabling efficient sampling without relying on Markov Chains, ultimately leading to superior performance.

\subsubsection{Hyperparameters for CE Attack}
\begin{figure}
    \centering
    \includegraphics[width=0.95\linewidth]{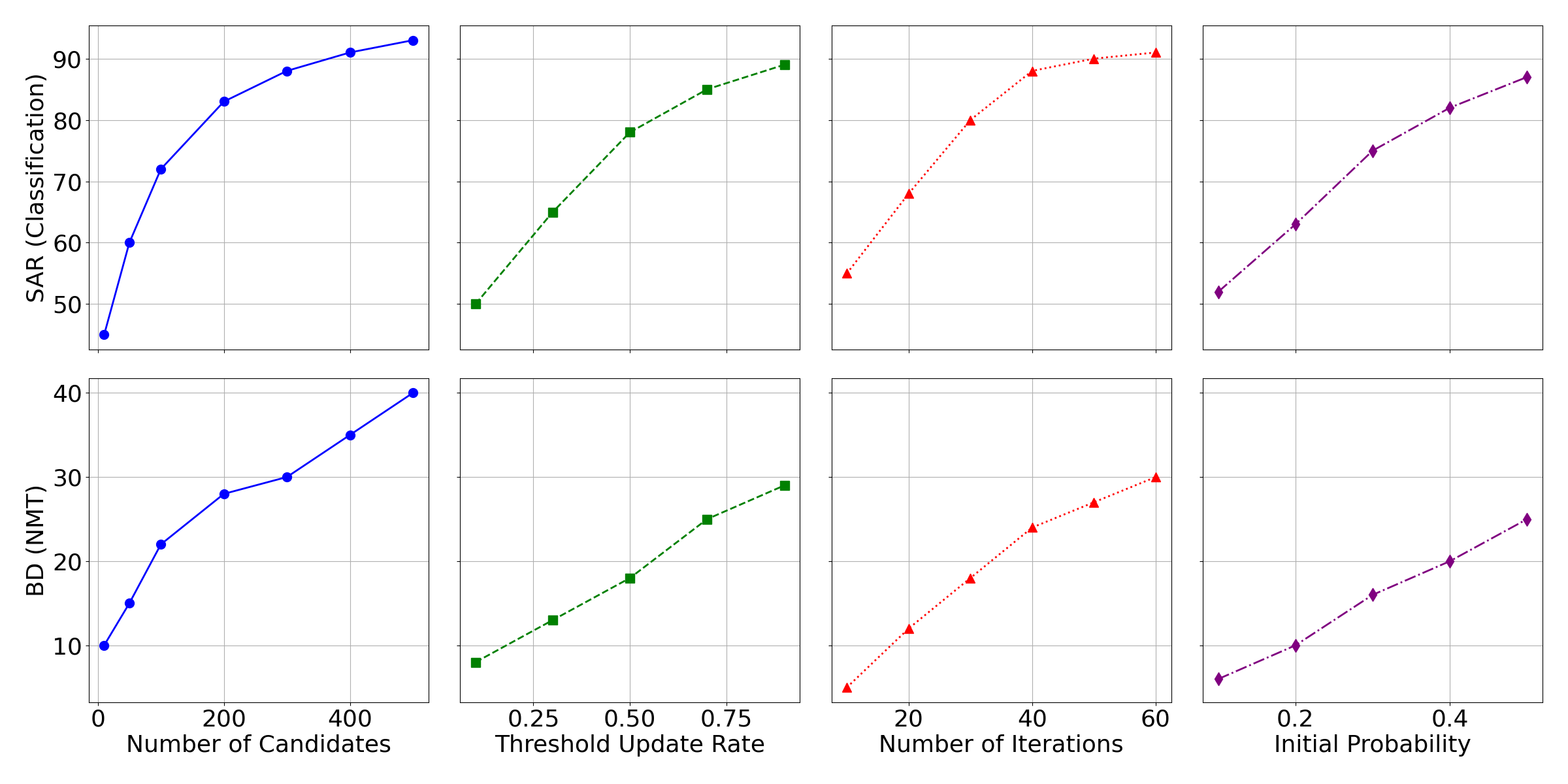}
    \caption{Ablation study results of CEA hyperparameters on SAR and BD metrics across classification and NMT tasks.}
    \label{fig: ablation hyperparameter}
\end{figure}

To ensure effective optimization in the Cross-Entropy Attack (CEA), we experimented with several key hyperparameters, each crucial to enhancing both attack efficiency and robustness. The experiments were conducted on IMDB for classification tasks and WMT T1 for neural machine translation (NMT), targeting BERT and mBART models, respectively. These experiments, visualized in Figure \ref{fig: ablation hyperparameter}, show the impact of each hyperparameter on attack success.

\paragraph{Number of Adversarial Candidates (N)}
The number of adversarial candidates \(N\) sampled in each CEA iteration defines the extent of perturbation exploration for each input. We varied \(N\) from 10 to 500 to understand its effect. As shown in Figure \ref{fig: ablation hyperparameter} (top left and bottom left), increasing \(N\) improves the Success Attack Rate (SAR) and BLEU Drop (BD) up to a certain point, after which gains taper off. This trend suggests that a larger candidate pool enhances the likelihood of finding impactful adversarial examples, but with diminishing returns beyond a threshold. We found \(N = 100\) to offer an optimal balance of effectiveness and efficiency.

\paragraph{Threshold Update Rate (\(\rho\))}
The threshold update rate \(\rho\) controls the rate of performance threshold refinement, which affects the candidate selection process in subsequent iterations. Our tests ranged \(\rho\) from 0.1 to 0.9, revealing that moderate rates (e.g., \(\rho = 0.5\)) achieve the best results (Figure \ref{fig: ablation hyperparameter}, second from the left). Lower values slowed convergence, while higher rates overly restricted candidate exploration, reducing effectiveness. Thus, \(\rho = 0.5\) provides an ideal balance, ensuring steady threshold tightening without sacrificing candidate diversity.

\paragraph{Number of Iterations (T)}
The number of iterations \(T\) determines the number of refinement cycles in the optimization process. Varying \(T\) from 10 to 60, we observed increasing SAR and BD metrics with higher iterations, stabilizing around \(T = 50\) (Figure \ref{fig: ablation hyperparameter}, second from the right). Beyond this point, additional iterations yielded minimal improvements. Setting \(T\) to 50 ensures that the algorithm achieves sufficient convergence while maintaining computational feasibility.

\paragraph{Initial Probability }
The initial probability distribution \(\hat{p}_{(i,j)} = \frac{1}{n_i}\) allocates equal likelihoods to each candidate, ensuring an unbiased starting point for exploration. Testing with values from 0.1 to 0.5, we observed a consistent increase in SAR and BD as the initial probabilities became more uniformly distributed, peaking around \(\hat{p}_{(i,j)} = 0.3\) (Figure \ref{fig: ablation hyperparameter}, far right). This setup allows CEA to more effectively explore diverse substitutions before refining the probability distribution toward optimal candidates.

In summary, these results highlight the importance of tuning each hyperparameter to achieve an optimal trade-off between attack effectiveness and computational efficiency.

\subsection{Attacking Models with Defense Mechanism} \label{defense}
\begin{figure*}
    \centering
    \includegraphics[width=0.95\linewidth]{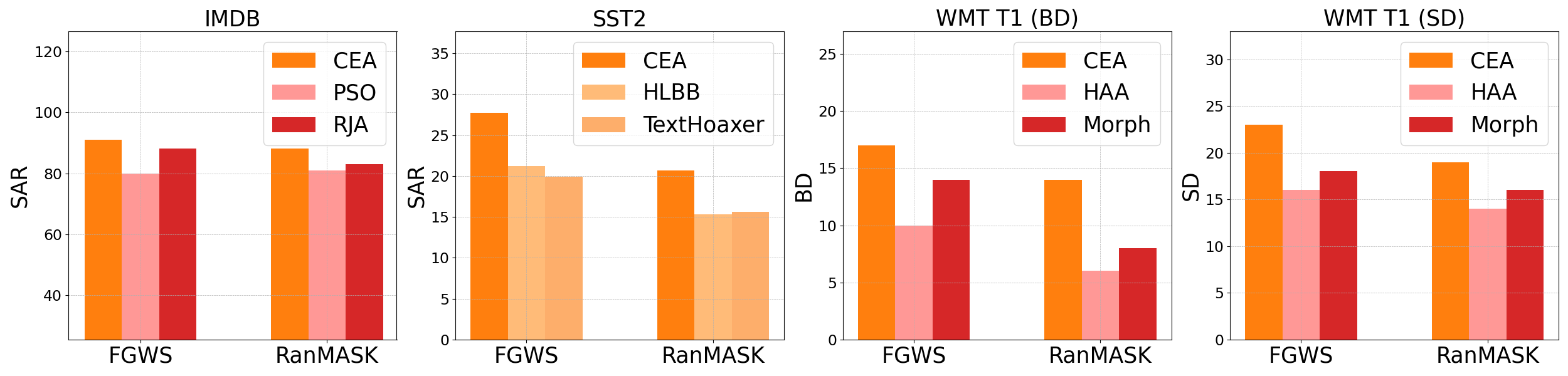}
    \caption{Performance comparison of adversarial attacks against BERT and T5 using two defense mechanisms, FGWS and RanMASK, across IMDB, SST2 and WMT T1 datasets.}
    \label{fig: defense}
\end{figure*}

Defending against textual adversarial attacks is crucial for the security of NLP models. We evaluated our attack methods using two defense strategies: Frequency-Guided Word Substitutions (FGWS)\cite{passive} for passive defense and Random Masking Training (RanMASK) \cite{active} for active defense. These defenses were tested on the BERT-C model using the IMDB and SST2 datasets, as shown in Figure \ref{fig: defense}. The results indicate that our method consistently outperforms the baselines, successfully bypassing the defense mechanisms and demonstrating the superior attacking performance of CEA.

\subsection{Transferability}\label{transfer}
\begin{figure}[t!]
    \centering
    \includegraphics[width=0.9\columnwidth]{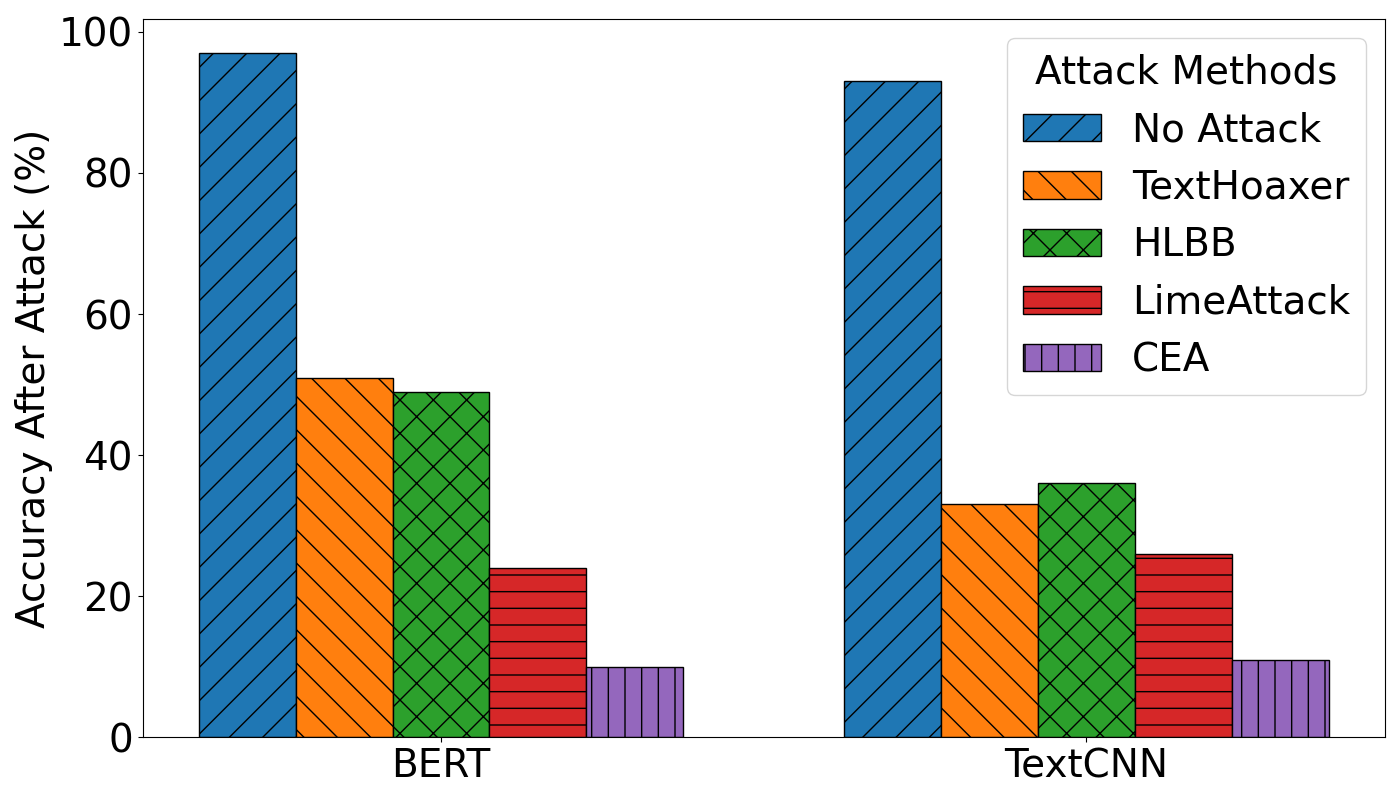}
    \caption{Performance of transfer hard-label attacks to victim models (BERT and TextCNN) on AG News. A lower accuracy of the victim models indicates a higher transfer ability (i.e., the lower, the better).}
    \label{fig: transfer}
\end{figure}

The transferability of adversarial examples refers to its ability to degrade the performance of other models to a certain extent when the examples are generated on a specific classifier \citep{Goodfellow2015fgsm}. To evaluate the transferability, we investigate further by exchanging the adversarial examples with hard-label attacks generated on BERT and TextCNN and the results are shown in Fig \ref{fig: adv acu}. When the adversarial examples generated by our methods are transferred to attack BERT and TexCNN, we can find that the attacking performance of CEA still achieves the best among baselines as illustrated in the Fig \ref{fig: transfer}. This suggests that the transferring attacking performance of the proposed methods consistently outperforms the baselines.

\subsection{Target Attack}\label{target}
A targeted attack seeks to manipulate a data sample with class \(y\) such that it is misclassified into a specified target class \(y^{\prime}\) while avoiding misclassification into other classes. The Cross-Entropy (CE) attack adapts seamlessly to this scenario by redefining its objective from minimizing \(1 - F_{y}(\mathbf{x})\) to maximizing \(F_{y^{\prime}}(\mathbf{x})\), enabling a direct focus on the target label. Targeted attack experiments conducted on the Emotion dataset demonstrate the superiority of CE attack over baselines such as PWWS, as shown in Table \ref{tab: target attack}. CE consistently achieves higher Success Attack Rate (SAR) while maintaining lower Modification Rate (Mod) and Grammar Error Rate (GErr) alongside improved Semantic Similarity (Sim) and Perplexity (PPL).

The superior performance of CE attack stems from its dynamic and adaptive optimization framework. By iteratively refining substitution probabilities, CE ensures that generated adversarial examples effectively misclassify into the target class with minimal perturbation, preserving fluency and semantic coherence. Unlike static methods, CE dynamically balances the trade-offs between attack effectiveness and imperceptibility, allowing it to handle diverse datasets and models robustly. This adaptability, coupled with its principled approach to optimizing adversarial objectives, ensures high success rates with fewer modifications and better overall quality of adversarial examples.

\begin{table}[t]
\caption{Targeted attack and imperceptibility-preserving performance on the Emotion dataset. The victim models are BERT-C and TextCNN classifiers. Results include both soft-label and hard-label attacks, comparing the proposed CEA method with baselines PWWS and HLBB. The statistics for better performance are vertically highlighted in bold.}
\centering
\label{tab: target attack}
\begin{tabular}{ccclllll}
\toprule
\multirow{2}{*}{Classifiers} & \multirow{2}{*}{Label Type} & \multirow{2}{*}{Attack Methods} & \multicolumn{5}{c}{Metrics} \\ \cmidrule(lr){4-8}
 &  &  & SAR$\uparrow$    & Mod$\downarrow$    & PPL$\downarrow$    & GE$\downarrow$    & SS$\uparrow$    \\ \midrule
\multirow{6}{*}{BERT-C} & \multirow{2}{*}{Soft-label} & PWWS       & 21.2            & 14.1              & 377                & 0.19              & 60              \\
                        &                             & \textbf{CEA} & \textbf{35.7}   & \textbf{8.7}      & \textbf{265}       & \textbf{0.11}     & \textbf{74}     \\ \cmidrule(lr){2-8}
                        & \multirow{2}{*}{Hard-label} & HLBB       & 26.9            & 11.4              & 305                & 0.17              & 65              \\
                        &                             & \textbf{CEA} & \textbf{34.2}   & \textbf{9.0}      & \textbf{275}       & \textbf{0.12}     & \textbf{71}     \\ \midrule
\multirow{6}{*}{TextCNN} & \multirow{2}{*}{Soft-label} & PWWS       & 32.6            & 11.1              & 345                & 0.22              & 63              \\
                         &                             & \textbf{CEA} & \textbf{63.2}   & \textbf{9.1}      & \textbf{240}       & \textbf{0.15}     & \textbf{70}     \\ \cmidrule(lr){2-8}
                         & \multirow{2}{*}{Hard-label} & HLBB       & 41.8            & 10.9              & 289                & 0.18              & 66              \\
                         &                             & \textbf{CEA} & \textbf{60.7}   & \textbf{9.3}      & \textbf{248}       & \textbf{0.16}     & \textbf{69}     \\ \bottomrule
\end{tabular}
\end{table}

\subsection{Adversarial Retraining}\label{retraining}
\begin{figure}[th]
    \centering
    \includegraphics[width=\columnwidth]{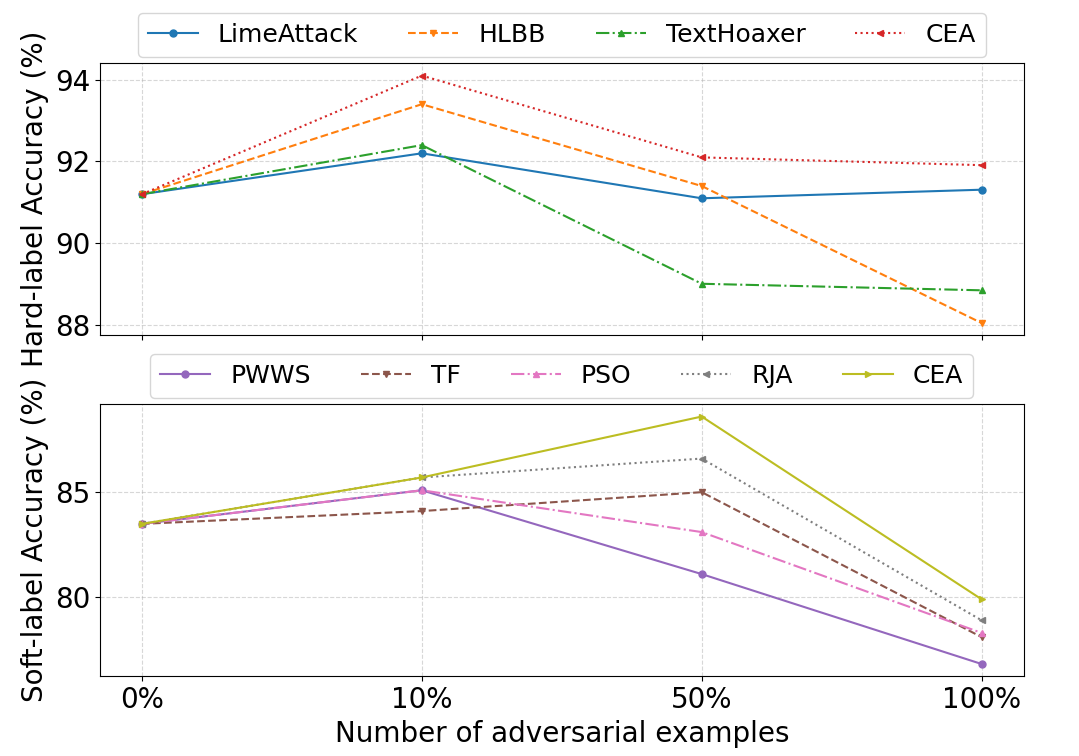}
    \caption{Performance of adversarially trained BERT with SST2 after joining different percentage of adversarial examples into the training set.}
    \label{fig: adv acu}
\end{figure}
\begin{figure*}
    \centering
    \includegraphics[width=\linewidth]{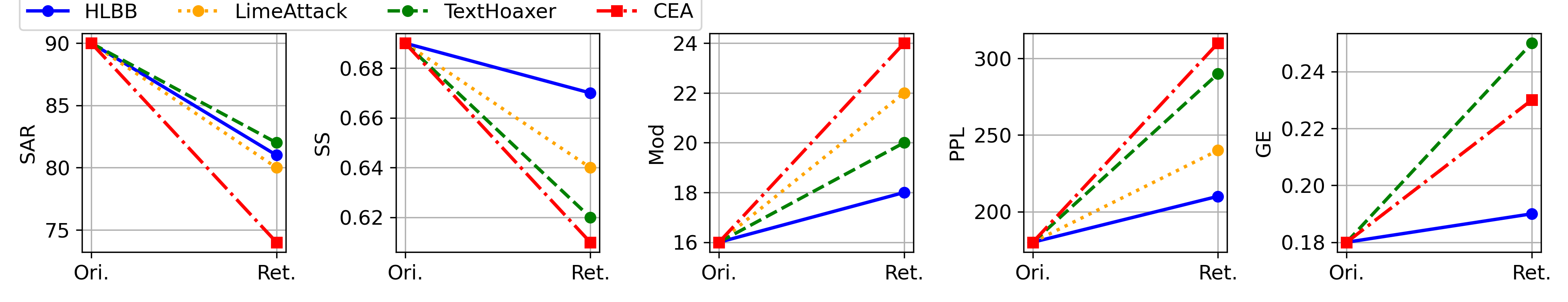}
    \caption{Rostbustness performance against adversarial attacks (CLARE) to retrained TextCNNs which joins 30\% adversarial examples from different hard-label attacking strategies to the training set of SST2.}
    \label{fig: adv robust}
\end{figure*}
Adversarial retraining has proven effective in enhancing both accuracy and robustness of machine learning models against adversarial attacks \cite{chivukula2020game}. We evaluated accuracy by incorporating {0\%, 10\%, 50\%, 100\%} adversarial examples into the SST2 training set for BERT classifiers. Figure \ref{fig: adv acu} shows that models retrained with CEA consistently outperform baselines, indicating better adaptation to the true distribution. Accuracy increases with more adversarial examples but declines beyond a certain point, aligning with existing literature \cite{li2021clare,rja}, suggesting that joining a proper amount of adversarial data will benefit the accuracy while excessive adversarial data can degrade performance.

For robustness, we use CEA to attack the classifiers trained with 30\% of adversarial examples from hard-label attack. As shown in Fig \ref{fig: adv robust}, adversarial training helps to decrease the attack success rate, imperceptibility and language quality, which means that the proposed CEA improves downstream models' robustness best compared with baselines.



\subsection{Platform and Efficiency Analysis}\label{efficiency}
Experiments were conducted on a RHEL 7.9 system with an Intel(R) Xeon(R) Gold 6238R CPU (2.2GHz, 28 cores - 26 enabled, 38.5MB L3 Cache), an NVIDIA Quadro RTX 5000 GPU (3072 Cores, 384 Tensor Cores, 16GB memory), and 88GB RAM. Table \ref{tab: CEA_efficiency} presents the time efficiency of different attack methods on the Emotion and IMDB datasets across soft-label, hard-label, and NMT attack types. CEA demonstrates consistently lower runtime, achieving 7.1 seconds per example for soft-label attacks on the Emotion dataset and 3.2 seconds on IMDB, outperforming baselines such as PWWS and HLBB.

These results underscore the computational efficiency of CEA while maintaining strong attack performance. Its ability to adaptively optimize adversarial examples without significant runtime overhead makes it a robust choice for both classification and NMT tasks, where efficiency and quality are critical.
\begin{table*}[t]
    \centering
    \caption{Time efficiency of attack algorithms for different datasets and attack types. Efficiency is measured in seconds per example, with lower values indicating better performance. The best performance for each scenario is \textbf{bold}.}
    \begin{tabular}{p{1.2cm}p{1.8cm}p{1cm}p{0.9cm}p{0.9cm}p{0.9cm}p{0.8cm}p{0.8cm}p{0.8cm}}
    \toprule
        Datasets & Type & PWWS & PSO & Lime.A & HLBB & HAA & Morph & CEA \\ \midrule
        \multirow{3}{*}{Emotion} 
        & Soft-label & 23.1 & 73.8 & - & - & - & - & \textbf{7.1} \\ 
        & Hard-label & - & - & 30.8 & 25.5 & - & - & \textbf{5.4} \\ 
        & NMT        & - & - & - & - & 52.4 & 60.2 & \textbf{4.3} \\ \midrule
        \multirow{3}{*}{IMDB} 
        & Soft-label & 3.7 & 166.9 & - & - & - & - & \textbf{3.2} \\ 
        & Hard-label & - & - & 4.9 & 92.3 & - & - & \textbf{4.8} \\ 
        & NMT        & - & - & - & - & 112.5 & \textbf{6.2} & \textbf{6.2} \\ 
        \bottomrule
    \end{tabular}
    \label{tab: CEA_efficiency}
\end{table*}

\section{Conclusion and Future Work}\label{conclusion}


Adversarial attacks have emerged as a critical challenge to the security and robustness of natural language processing (NLP) models, undermining their reliability in key applications such as sentiment analysis, machine translation, and content moderation \citep{szegedy2013intriguing, goodfellow2014explaining}. In response to these vulnerabilities, we introduced the Cross-Entropy Attack (CEA), a novel framework that leverages cross-entropy optimization to generate highly effective yet imperceptible adversarial examples \citep{papernot2016limitations, kurakin2016adversarial}. By formulating tailored optimization objectives for both classification and sequence-to-sequence tasks, CEA achieves state-of-the-art performance across diverse datasets and victim models \citep{ebrahimi2017hotflip, alzantot2018generating}. Our extensive empirical evaluation demonstrates CEA's adaptability, achieving superior success rates, semantic preservation, and linguistic fluency when compared to established baselines \citep{liang2017deep, zhao2017generating}.

While adversarial examples highlight vulnerabilities in NLP systems, they also serve as a valuable mechanism for testing and enhancing model robustness \citep{madry2017towards, sinha2017certifying}. Through adversarial retraining, we demonstrated that integrating adversarial examples into the training process significantly improves model resilience \citep{goodfellow2014explaining, kurakin2016adversarial}. However, our findings also caution against excessive reliance on such examples, which may inadvertently degrade performance on clean data \citep{tsipras2018there, zhang2019theoretically}. These results underscore the need for balanced approaches to adversarial training, ensuring robust defenses while maintaining generalization capabilities.

The prevalence of deep neural networks and large pre-trained models underscores the necessity for proactive defenses against adversarial threats \citep{devlin2018bert, radford2018improving}. Our future research will focus on designing adaptive defense mechanisms capable of mitigating emerging attack techniques while preserving model interpretability and usability \citep{xu2017virginia, jia2019certified}. Additionally, exploring the application of cross-entropy optimization principles to develop more robust and secure NLP systems offers a promising direction \citep{papernot2016limitations, kurakin2016adversarial}. 
\backmatter








\section*{Declarations}
\begin{itemize}
\item Funding: Not Applicable.
\item Competing interests: Not Applicable.
\item Ethics approval: Not Applicable.
\item Consent to participate: The authors give their consent to participate.
\item Consent for publication: The authors give their consent to the publication of all information in this paper.
\item Availability of data and materials: All of the datasets are available on Huggingface (\url{https://huggingface.co/datasets}) and on our GitHub site (\url{https://github.com/MingzeLucasNi/RCEAgit})
\item Code availability: All codes from our experiments are available at \url{https://github.com/MingzeLucasNi/CEA.git}
\item Authors' contributions: 
 Mingze Ni contributed to conceptualization, theoretical analysis, experiments and draft preparation; Wei Liu contributed to conceptualization, theoretical analysis, draft writing and editing.
\end{itemize}









\bibliography{sn-bibliography}

\end{document}